\documentclass[journal,twoside]{IEEEtran}
\pdfoutput=1
\usepackage{epsfig}
\usepackage{graphicx}
\usepackage{subfigure}
\usepackage{multirow}
\usepackage{algorithm}
\usepackage{algorithmic}
\usepackage{cite}
\usepackage{CJK}
\usepackage{amsmath}
\usepackage{makecell}
\makeatletter
\newcommand{\Rmnum}[1]{\expandafter\@slowromancap\romannumeral #1@}

\ifCLASSINFOpdf

\else

\fi
\begin{document}
%
\title{An Intelligent CNN-VAE Text Representation Technology Based on Text Semantics for Comprehensive Big Data
}


\author{\IEEEauthorblockN{Genggeng Liu\IEEEauthorrefmark{1},
Canyang Guo\IEEEauthorrefmark{1},
Lin Xie\IEEEauthorrefmark{1},
Wenxi Liu\IEEEauthorrefmark{1},
Naixue Xiong\IEEEauthorrefmark{2}, and
Guolong Chen\IEEEauthorrefmark{1}}\\
\IEEEauthorblockA{\IEEEauthorrefmark{1}College of Mathematics and Computer Science, Fuzhou University, Fuzhou, 350100 China}\\
 (E-mail: liu\_genggeng@126.com; canyangguo@163.com; xlin540@163.com; wenxi.liu@hotmail.com; cgl@fzu.edu.cn)\\
\IEEEauthorblockA{\IEEEauthorrefmark{2}Department of Business and Computer Science, Southwestern Oklahoma State University, Weatherford, OK 73096 USA\\
 (E-mail: neal.xiong@swosu.edu).}
}

%



\IEEEtitleabstractindextext{%
\begin{abstract}
In the era of big data, a large number of text data generated by the Internet has given birth to a variety of text representation methods. In natural language processing (NLP), text representation transforms text into vectors that can be processed by computer without losing the original semantic information. However, these methods are difficult to effectively extract the semantic features among words and distinguish polysemy in language. Therefore, a text feature representation model based on convolutional neural network (CNN) and variational autoencoder (VAE) is proposed to extract the text features and apply the obtained text feature representation on the text classification tasks. CNN is used to extract the features of text vector to get the semantics among words and VAE is introduced to make the text feature space more consistent with Gaussian distribution. In addition, the output of the improved word2vec model is employed as the input of the proposed model to distinguish different meanings of the same word in different contexts. The experimental results show that the proposed model outperforms in k-nearest neighbor (KNN), random forest (RF) and support vector machine (SVM) classification algorithms.
\end{abstract}

\begin{IEEEkeywords}
Natural language processing(NLP), text representation, variational autoencoder(VAE), convolutional neural network(CNN), feature extraction, polysemy.
\end{IEEEkeywords}}

\maketitle

\IEEEdisplaynontitleabstractindextext

%
\IEEEpeerreviewmaketitle

\section{Introduction}
%
%
%
%
\IEEEPARstart{I}{n} recent years, with the rapid development of Internet technology, the Internet generates extensive data every day including text, image, video, audio, etc. Text data plays a significant role because it not only occupies a large part of the Internet data but also can be applied in many real world scenarios including getting the current hotspots, developing question answering system and machine translation. Taking search engine as an example, there are tens of millions of search task requests every day, most of which use text information as the input of search task [1].

As the basic task of natural language processing (NLP), text representation transforms unstructured natural language into a structured form that contains unique
semantic information of the original text data and thus can be processed and analyzed by computer [2]. According to the types of natural language, text representation can be categorized into four different granularity representations, word representation, sentence representation, paragraph representation and document representation. For different granularity of text representation, they share the same purpose of extracting the most important semantic information in different applications [3]. In traditional text representation method, bag-of-words models model which considered every word in the dictionary as a feature of text representation have been widely used [4]. A document is composed of text vector features corresponding to all words in the dictionary. If a word is included in the document, the corresponding value in the text vector is 1, otherwise it is 0. Considering that the bag-of-words models uses all the words, the dimension of the final text feature vector will lead to dimension disaster with the increase of data scale, which takes up a lot of computation costs and running time. Therefore, a reasonable text representation method is necessary in NLP and feature extraction method is an effective way to reduce the dimension of text feature vector [5].

 The traditional method of word vector representation is one-hot. The dimension of word vector representation is the same as the number of all different words in the corpus. The vector position corresponding to words is set to 1 and the rest is set to 0. For a corpus composed of $n$ different words, the dimension of each word vector $n$. The dimension of the word vector generated by this method results in the high dimension and sparsity for the word vector is directly proportional to the size of the corpus, which makes it difficult to obtain semantic information. Moreover, it is unable to calculate the correlation between near-synonyms by coding in this way.  For example, the words "pleased" and "happy" are assumed to be the word vectors obtained by one-hot coding as [1,0,0,0] and [0,1,0,0]. "Pleased" and "happy" have similar meanings, but the cosine similarity of the two word vectors is 0. Obviously the similarity between the word vectors calculated by the method is unreasonable and their semantic information cannot be reflected.
The development of computer processing power promotes the application of neural network in NLP to a certain extent. In order to solve the problem of word vector, the researchers proposed the word embedding based on neural network language model (i.e. distributed representation) [6]. The early proposed neural network language model used the current word to predict the next word to build a neural network model from the perspective of probability. This model is mainly for the prediction task of experimental words and the word vector is only the by-product of the optimized model. Thereafter, how to use neural network to train word vector has been studied and improved by researchers.
Huang et al. [7] held that it should not only focus on the relationship of local context, but also use the full text information to assist the training of local word information. Therefore, the word vector contains more semantic information and the representation of polysemy can also be solved by the joint representation of multiple word vectors.
Mikolov et al. [8] proposed word vector training model word to vector (word2vector), which aims to transform text into word vector through neural network and has been widely studied and used by researchers. There are four categories of text representation methods based on neural network according to different granularity (i.e. words, sentences, paragraphs and documents). In order to solve the problems of data sparsity, dimension disaster and lack of semantic expression in the bag-of-words model used in sentences and documents, the multi-layer neural network is used to map and extract features. Ji et al. [9] proposed another word vector training model Wordrank which calculate the similarity of words and other different standards to get word vector representation and is clever at representation of similar words . Considering the complexity of the information interaction platform in the Internet, Dhingra et al. [10] proposes the tweet2vec model, which is a distributed representation model based on character combination in the complex social media environment. Hill et al. [11] considered that there is an optimal representation method for different application tasks and proposed a sentence level text representation method to learning from unlabeled text data. It can not only optimize the training time, but also improve the portability by employing unsupervised method. Le et al. [12] proposes two levels of text representation models including sentence and document, both of which are based on the word2vec model. Kalchbrenner et al. [13] proposed a text representation model based on dynamic convolution neural network which adopts dynamic pooling technology and is expert in emotion recognition. Considering the advantages of convolution neural network (CNN) in extracting local features, Hu et al. [14] proposed employing CNN to extract semantic information among words in sentences and achieved excellent results in sentence matching tasks. On this basis, Yin et al. [15]proposes Bi-CNN-MI text representation model which can extract four different granularity text representations from sentences and realize the interaction of these four different granularity features through the CNN so as to adapt to the synonymous sentence detection task. Zhang et al. [16] proposes a text representation model based on dependency sensitive CNN at the sentence and document level, which applies persisting sensitive information and feature extracting on inputting word vectors.

This paper proposes a text feature representation model based on CNN and variational autoencoder (VAE), which is a feature representation method from word vector to text vector. The main advantages of the proposed methods are its lower computation cost, extracting semantic information and distinguishing polysemy. In this method, CNN is used to extract the features of text vector as text representation to get the semantics among words and VAE is introduced to make the text feature space more consistent with Gaussian distribution. In addition, considering the polysemy of words, this paper uses the output of the improved word2vec model as the input of the proposed model to distinguish different meanings of the same word in different contexts. A word is mapped into multiple vectors responding to different topics by this model  to solve  polysemy. Four evaluation metrics, including accuracy, recall, precision and F-score, are used to evaluate the performance of the proposed model. Experimental results show that the model has better performance than w2v-avg and CNN-AE in k-nearest neighbor (KNN), random forest (RF) and support vector machine (SVM) classification.

The remainder of this paper are organized as follows. Section 2 and Section 3 introduce the related algorithm and proposed CNN-VAE model based on text semantics. Section 4 illustrates a case study of open dataset Cnews. Section 5 shows the conclusion and future research directions.
\section{Related Works}
This section introduces the related work of text representation.  Section 2.1 introduces the classic word2vec model including CBOW and skip-gram model. Section 2.2 shows the text representation method based on word2vec and Section 2.3 illustrates CNN algorithm and Section 2.4 introduces the feature extracting method based neural network.
\subsection{Word2vec}
Word2vec algorithm proposed by Mikolov aims to obtain the text word vector model by training a neural network to get the weight matrix of the network [8]. Word2vec model includes CBOW and Skip-gram training models and they only contain simple neural network structures: input layer, projection layer and output layer. CBOW and skip-gram realize the network based on different conditions: CBOW predicts the probability of the central word through the context word while skip-gram model predicts the probability of the context word through the central word [17].
\begin{figure}[!t]
\centering
\includegraphics[width=3.4in]{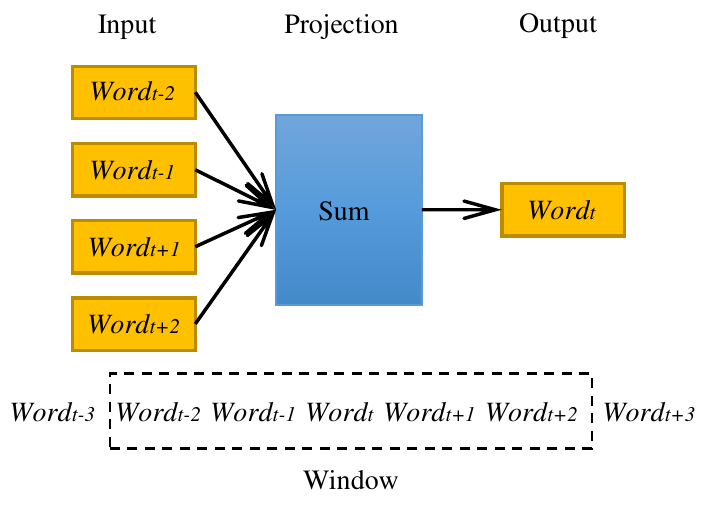}
\caption{ The structures of CBOW model.}
\label{fig_sim}
\end{figure}
\subsubsection{CBOW Based Method}
The structures of CBOW model is shown in Fig. 1, the input is the context $Context\left( {{w_t}} \right)$ of the central word ${w_t}$ in the sliding window and the object is to predict the central word ${w_t}$. The output of CBOW model is a softmax classifier function, which is usually implemented by using negative sampling method. The central word ${w_t}$ in the sliding window is judged as a positive sample and other words are judged as negative samples. In the output layer, there is a negative sample set $NEG({w_t})$ for the central word ${w_t}$ so that any word $u$ in the corpus meets the Equation (1).
\begin{equation}
\label{eqn_example}
{R^{{w_t}}}(u) = \{ \begin{array}{*{20}{c}}
{\;1,\;\;u = {w_t}}\\
{0,\;\;u \ne {w_t}}
\end{array}
\end{equation}
where ${R^{{w_t}}}(u)$ determines whether $u$ is a positive sample.
For the central word and its context $\langle {w_t},Context{\rm{ (}}{w_t})\rangle $, CBOW model needs to maximize the objective function $g({w_t})$ shown in Equation (2).
\begin{equation}
\label{eqn_example}
	g({w_t}) = \prod\nolimits_{u \in (\{ {w_t}\}  \cup NEG({w_t})} {p(u|Context{\rm{ (}}{w_t}))}
\end{equation}
The formal of $p\left( {u|Context\left( {{w_t}} \right)} \right)$  is shown in Equation (3).
\begin{equation}
\label{eqn_example}
p(u|Context{\rm{ (}}{w_t})) = \{ _{1 - \sigma (X_{{w_t}}^T{\theta ^{{w_t}}}),{R^{{w_t}}}(u) = 0}^{\sigma (X_{{w_t}}^T{\theta ^{{w_t}}}),\;\;\;\;{R^{{w_t}}}(u) = 1}	
\end{equation}
where ${X_{{w_t}}}$ stands for the sum of the word vectors of the words in the context $Context{\rm{ (}}{w_t})$ . The objective function can be converted to Equation (4).
\begin{equation}
\label{eqn_example}
g({w_t}) = \sigma (X_{{w_t}}^T{\theta ^{{w_t}}})\prod\nolimits_{u \in NEG({w_t})} {[1 - \sigma (X_{{w_t}}^T{\theta ^u})]}
\end{equation}	
where  $\sigma (X_{{w_t}}^T{\theta ^{{w_t}}})$ represents the prediction probability of getting central word $w_t$  via CBOW when the context is $Context{\rm{ (}}{w_t})$ .  $\sigma (X_{{w_t}}^T{\theta ^u})$is the prediction probability of getting central word $u$  when the context is $Context{\rm{ (}}{w_t})$. In order to maximize  $g({w_t})$, it is necessary to maximize $\sigma (X_{{w_t}}^T{\theta ^{{w_t}}})$ to increase the probability of positive samples. At the same time, minimize all the $\sigma (X_{{w_t}}^T{\theta ^u})$ to reduce the probability of negative samples.
\begin{figure}[!t]
\centering
\includegraphics[width=3.4in]{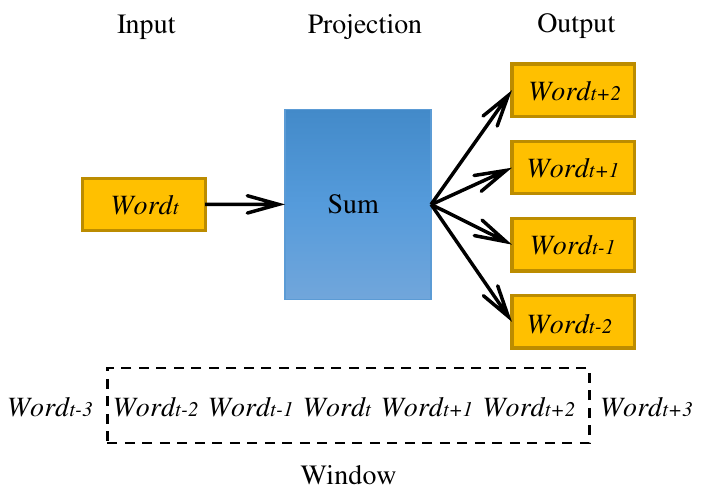}
\caption{ The structures of Skip-gram model.}
\label{fig_sim}
\end{figure}
\subsubsection{Skip-gram Based Method}
The structures of Skip-gram model is shown in Fig. 2. In contrast to CBOW model, Skip-gram determines the central word $w_i$  and then predicts the context $Context{\rm{ (}}{w_t})$ in its sliding window through the central word. The output of skip-gram model corresponds to a positive sample $u \in Context{\rm{ (}}{w_t})$. On the contrary,  $NEG(u)$ represents negative sample set which do not belong to the positive sample$Context{\rm{ (}}{w_t})$ . Therefore, the data $\langle {w_t},Context{\rm{ (}}{w_t})\rangle $ in the sliding window meets the objective function shown in Equation (5).
\begin{equation}
\label{eqn_example}
	g({w_t}) = \prod\nolimits_{u \in Context({w_t})} {\prod\nolimits_{x \in \{ u\}  \cup  \in NEG(u)} {p(x|{w_t})} }
\end{equation}	

Similar to CBOW method, the relation${R^u}(x)$ is introduced to determine whether $x$ is a positive sample. When it is a positive sample, $x=u$ and ${R^u}(x)=1$ . When it is a negative sample,  $x=NEG(u)$ and ${R^u}(x)=0$. The conditional probability is shown in Equation (6).
\begin{equation}
\label{eqn_example}
	p(x|{w_t}) = {[\sigma (v{({w_t})^T}{v_x})]^{{R^u}(x)}} \cdot {[1 - \sigma (v{({w_t})^T}{v_x})]^{1 - {R^u}(x)}}
\end{equation}	
where  $\sigma (v{({w_t})^T}{v_x})$ represents the prediction probability of getting context $context({w_t})$  when the input of central word is ${w_t}$ .  $v({w_t})$ is the relationship between input layer and hidden layer.  ${v_x}$ is the output of network. For maximizing  $g({w_t})$, it is necessary to maximize $\sigma (v{({w_t})^T}{v_x})$ when positive sampling and minimize the $\sigma (v{({w_t})^T}{v_x})$ when negative sampling.
\subsection{Convolution Neural Network}
CNN is a deep artificial neural network including convolution calculation and polling calculation. It has excellent performance in the field of computer vision and NLP with its powerful feature extraction ability [18]. In the field of NLP, CNN employed convolution kernel to convolute text matrix of different length and the vectors through convolution kernels are calculated by pooling layer to extract features for text classification [19].
\subsubsection{Convolution Layer}
Convolution layer can extract local features by convolution kernel and get the final output by activation function. For $k\-th$ convolution kernel, the convolution process is shown in Equation (7).
\begin{equation}
\label{eqn_example}
{c_i} = f\left( {{w_k} * x + {b_k}} \right)
\end{equation}	
where ${w_k}$ and ${b_k}$ are the weight matrix and bias of $k\-th$ convolution kernel. $x$ stands for the input matrix. $f$ represents the activation function. For a sentence of length $n$ , its feature vector is shown in Equation (8).
\begin{equation}
\label{eqn_example}
	c = [{c_1},{c_2},{c_3}...{c_n}]
\end{equation}	
\subsubsection{Pooling Layer and Full Connection Layer}
In the pooling layer of the model, the maximum pooling technology is used to extract the feature value $C$ , which contains the highest semantic information in the local features of the convoluted window, as is shown in Equation (9).
\begin{equation}
\label{eqn_example}
C = \max (c)
\end{equation}
The complexity of the parameters in the convolutional neural network can be effectively reduced by using max pooling. For a window with  $k$ convolution kernels, the feature vectors obtained are shown in Equation (10).
\begin{equation}
\label{eqn_example}
\hat C = [{\hat C_1},{\hat C_2},{\hat C_3}...{\hat C_n}]
\end{equation}	
Next, the activation function is used to predict the labels of sentences, as is shown in Equation (11).
\begin{equation}
\label{eqn_example}
y = f(\hat C)	
\end{equation}	
where $y$ is the predicted label ($1$ stands for positive label and $0$ stands for negative label).  The loss function can be expressed as Equation (12).
	\begin{equation}
\label{eqn_example}
loss = \frac{1}{{2n}}\sum\limits_{i = 1}^n {{{\left\| {{y_i} - {{\hat y}_i}} \right\|}^2}}
\end{equation}	
where $y_i$ is the actual label. The loss function can be used to learn the parameters of the network via iteration method.
\subsection{Autoencoder}
Considering the high dimension and sparsity of the word vector generated by traditional methods, the feature selection method based on neural network is proposed to reduce the feature space of the text vector. Autoencoder (AE) can be regarded as a special neural network including input layer, hidden layer and output layer [20]. The number of neurons in the output layer is equal to the number of neurons in the input layer. As an unsupervised feature learning method, the purpose of AE is to obtain hidden feature and remove redundant information.
The forward propagation process of AE is similar to that of traditional neural network, which will not be discussed here. The aim of AE training is to minimize the reconstruction error. The loss function is shown in Equation (13).
\begin{equation}
\label{eqn_example}
L = \sum\limits_{i = 1}^n {{{\left\| {{x_i} - {y_i}} \right\|}^2}}
\end{equation}	
where $x_i$  and $y_i$  represent input data and output data.  $L$ is the loss function of AE and $n$ is the number of data.
\begin{figure*}[!t]
\centering
\includegraphics[width=6in]{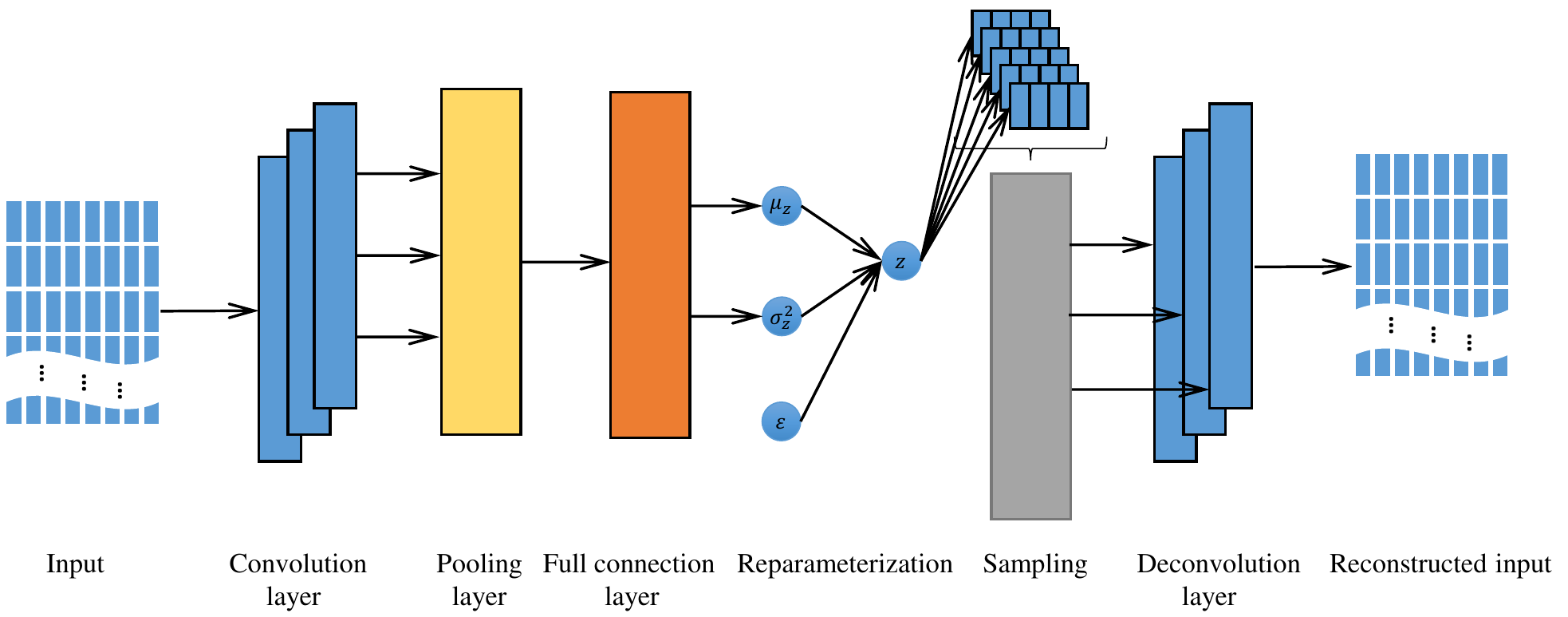}
\caption{ The structures of CNN-VAE model.}
\label{fig_sim}
\end{figure*}
\section{Text Feature Representation Model Based on CNN-VAE}
This paper combine the network framework of the VAE with the CNN to extract the text feature representation from the word vector. The traditional AE employs full connection layer, which is replaced by CNN for it can learn local features efficiently. Moreover, VAE makes the vector feature space conform to the function of Gaussian distribution, which makes the final text feature representation richer in semantic information. The structure of CNN-VAE is shown in Fig. 3.
The CNN is used to realize the network structure of VAE, because CNN can learn better local features from the input matrix. In this part, CNN network is used to build the VAE network framework, so that CNN combines the VAE feature extraction and the function of making the vector feature space conform to the Gaussian distribution in its own text feature extraction, making the final text feature representation richer in semantic information.
Similar to AE, the proposed method includes decoding and encoding. The decoding process can be regarded as a general CNN which can achieve the purpose of feature extraction via convolution and pooling. A matrix  $x$ can be obtained by splicing the word vectors corresponding to the words appearing in an article is putted into the model. Random number  $z$ corresponding to Gaussian distribution is generated according to the mean value $\mu $  and variance $\sigma $  of the output of the convolution encoder. Suppose that there is a set of functions ${p_\theta }(x|z)$  for generating  $x$ from $z$ , each of which is uniquely determined by $\theta $ . The goal of the VAE is to maximize ${p_\theta }(x)$  by optimizing  $\theta $ to make the generated data  $\hat x$ similar to original data  $x$ as possible and the formula is shown Equation (14).
\begin{equation}
\label{eqn_example}
{p_\theta }(x) = \int {{p_\theta }} \left( {x|z} \right){p_\theta }\left( z \right)dz
\end{equation}	
In order to obtain ${p_\theta }(z)$, the encoder network ${p_\theta }(z|x)$  is introduced. An approximate posterior ${q_\varphi }(z|x)$ obeying Gaussian distribution is employed to take the place of ${p_\theta }(z|x)$ for the latter is difficult to obtain by calculation. Kullback Leibler (KL) divergence is applied to measure the similarity between two distributions [21] and the formula is shown in Equation (15).
\begin{equation}
\label{eqn_example}
	\begin{array}{l}
{D_{KL}}({q_\varphi }(z|x)\parallel {p_\theta }(z|x))\\
 = \sum {{q_\varphi }(z|x)} \left[ {\log {q_\varphi }(z|x) - \log {p_\theta }(x|z) - \log {p_\theta }(z)} \right] + \log {p_\theta }(x)
\end{array}
\end{equation}	
Therefore, the formal of $\log {p_\theta }(x)$ can be expressed as Equation (16).
\begin{equation}
\label{eqn_example}
\begin{array}{l}
\log {p_\theta }(x) = {D_{KL}}({q_\varphi }(z|x)\parallel {p_\theta }(z|x))\\
 - \sum {{q_\varphi }(z|x)} \left[ {\log {q_\varphi }(z|x) - \log {p_\theta }(x|z) - \log {p_\theta }(z)} \right]
\end{array}
\end{equation}	
Considering the KL divergence is not negative, the loss function can be expressed as Equation (17).
\begin{equation}
\label{eqn_example}
\begin{array}{l}
L\left( {\theta ,\varphi ;x} \right) \\
=  - \sum {{q_\varphi }(z|x)} \left[ {\log {q_\varphi }(z|x) - \log {p_\theta }(x|z) - \log {p_\theta }(z)} \right]\\
=  - {D_{KL}}({q_\varphi }(z|x)\parallel {p_\theta }(z|x)) + \sum {{q_\varphi }(z|x)} [\log {p_\theta }(x|z)]
\end{array}
\end{equation}	
where $L\left( {\theta ,\varphi ;x} \right)$  is loss function.  $ - {D_{KL}}({q_\varphi }(z|x)\parallel {p_\theta }(z|x))$ is regularizer and $\sum {{q_\varphi }(z|x)} [\log {p_\theta }(x|z)]$ is reconstruction error. Considering  ${p_\theta }(z)$ obeys the Gaussian distribution $N{\rm{ }}\left( {0;I} \right)$ and ${q_\varphi }(z|x)$  obeys the Gaussian distribution $n{\rm{ }}\left( {\mu ;{\sigma ^2}} \right)$ so that the regularizer can be expressed as Equation (18).
\begin{equation}
\label{eqn_example}
	 - {D_{KL}}({q_\varphi }(z|x\parallel {p_\theta }(z)) = \frac{1}{2}\sum {(1 + \log ({\sigma ^2} - {\mu ^2} - {\sigma ^2})} )
\end{equation}	
where $j$  is the dimension of $z$ . Monte Carlo evaluation is used to solve the reconstruction error shown as equation (19).
\begin{equation}
\label{eqn_example}
	\sum {{q_\varphi }(z|x)[\log {p_\theta }(x|z)]}  = \log {p_\theta }(x|z)
\end{equation}	

The technique of reparameterization is employed considering $z$ is not derivable. In this way, the original derivation of $z$ can be converted into the derivation of $\mu $  and $\sigma $ shown as Equation (20).
\begin{equation}
\label{eqn_example}
z = \mu  + \varepsilon  \cdot \sigma
\end{equation}	
$\varepsilon$ can be considered as sampling from $N(0,1)$ and $\log {p_\theta }(x|z)$  can be expressed as Equation (21).
\begin{equation}
\label{eqn_example}
	\log {p_\theta }\left( {x|z} \right) =  - \sum {\left( {\left( {\frac{1}{2}\left\| {\frac{{x - \mu }}{\sigma }} \right\|} \right) + \log \left( {\sqrt {x\pi } \sigma } \right)} \right)}
\end{equation}	
In conclusion, the loss function can be expressed as Equation (22).
\begin{gather}
\label{eqn_example}
	L\left( {\theta ,\varphi ;x} \right) = \frac{1}{2}\sum {(1 + \log ({\sigma ^2} - {\mu ^2} - {\sigma ^2})} ) \notag\\
- \sum {\left( {\left( {\frac{1}{2}\left\| {\frac{{x - \mu }}{\sigma }} \right\|} \right) + \log \left( {\sqrt {x\pi } \sigma } \right)} \right)}
\end{gather}	
\begin{table}
\makeatletter\def\@captype{table}\makeatother
\caption{Parameter Value.}
\label{tab:1}       
\centering
\setlength{\tabcolsep}{7mm}
\begin{tabular}{ c  c }
\hline\noalign{\smallskip}
Parameter	&setting\\
\noalign{\smallskip}\hline\noalign{\smallskip}
Dimension of word vector	&128\\
Number of iterations	&30\\
Dropout	&0.5\\
Size of hidden layer $z$	&128\\
Learning rate	&0.001\\
Padding(Maximum words per article)	&100\\
\noalign{\smallskip}\hline
\end{tabular}
\end{table}
The training process of the network is divided into the following steps. Firstly, the data of training set is used to find the optimal parameters of the model when the loss function is minimized. Next, when the network loss converges, the verification set is inputted to the encoder network obtained by training set to get the corresponding text representation vector. The corresponding classification accuracy is obtained by inputting the text representation vector into the classification model. Finally, the learning rate is adjusted many times to obtain the optimal parameters of the model. After several adjustments, the optimal model parameters are shown as Table I.
\section{CNN-VAE text representation model based on text semantics}
Polysemy means that the same word can express multiple meanings, which is the general ambiguity in natural language.
Word vectors generated by word2vec correspond to only one word in the corpus, which ignores the ambiguity in natural language.
Therefore, this paper proposes a CNN-VAE text representation model based on text semantics. A word is mapped into multiple vectors responding to different topics by this model  to solve  polysemy. In this model, word vector generated by word2vec on the topic model is used as the input of CNN-VAE model.
\subsection{LDA}
In 2003, Kim et al. [22] proposed the latent Dirichlet allocation model (LDA),  a topic model, can extract the topic information from an article. The hypothetical structure of LDA is that an article is composed of different topics and each topic is composed of different words, that is "document-topic-word" structure. LDA is a three-layer Bayesian probability model based on this structure.
\begin{figure}[!t]
\centering
\includegraphics[width=3in]{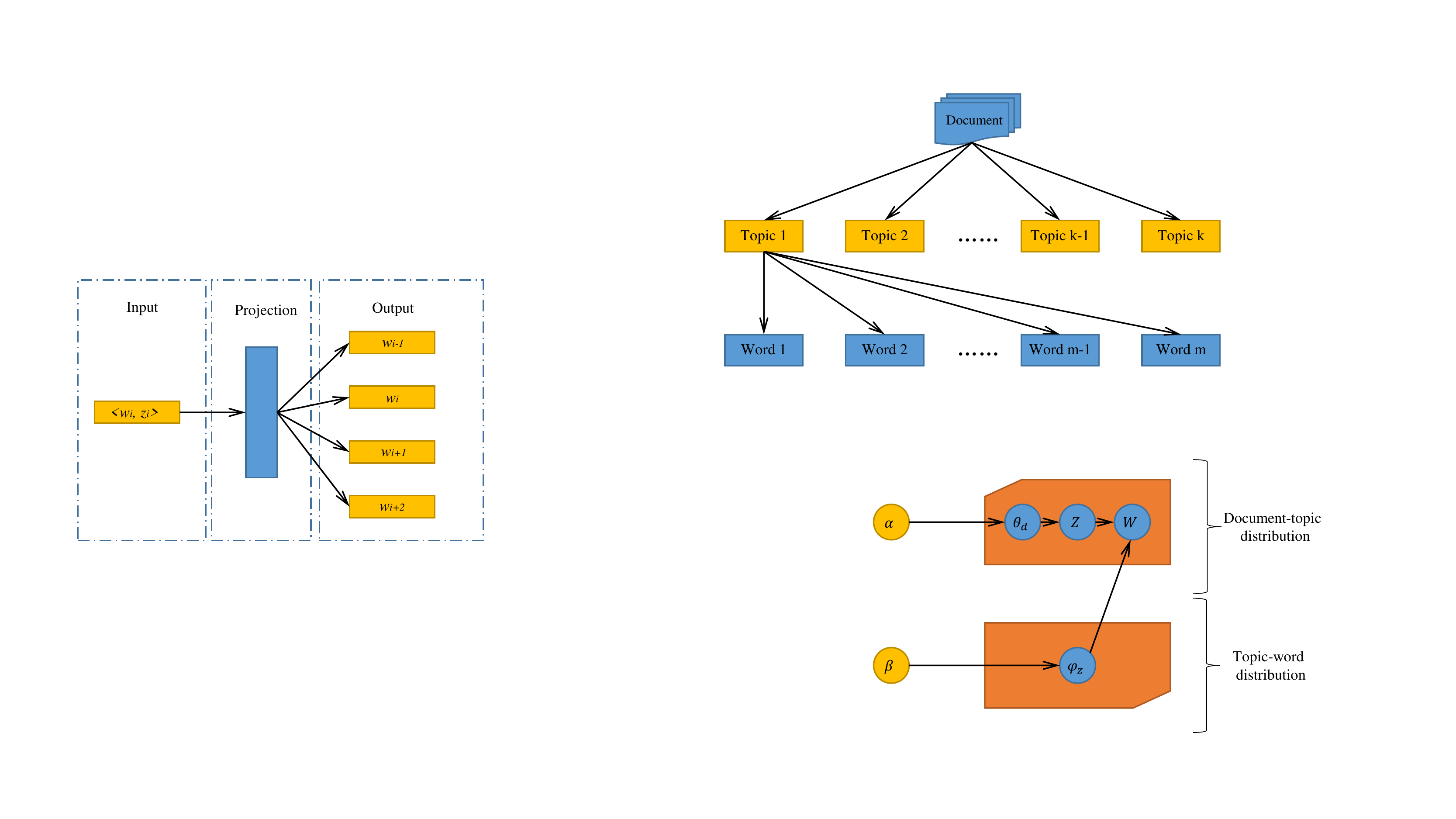}
\caption{ The structures LDA model.}
\label{fig_sim}
\end{figure}
The principle of LDA model is shown in Fig. 4. Suppose that there are $k$ topics in document set $D$ and each document is made up of these $k$ topics according to different probabilities, which stores the matrix of corresponding probabilities (i.e. document topic matrix).
 Similarly, each topic contains $m$ words and each topic is also made up of $m$ words according to different probabilities, which stores the moments of corresponding probabilities (i.e. topic vocabulary matrix).
\begin{figure}[!t]
\centering
\includegraphics[width=3in]{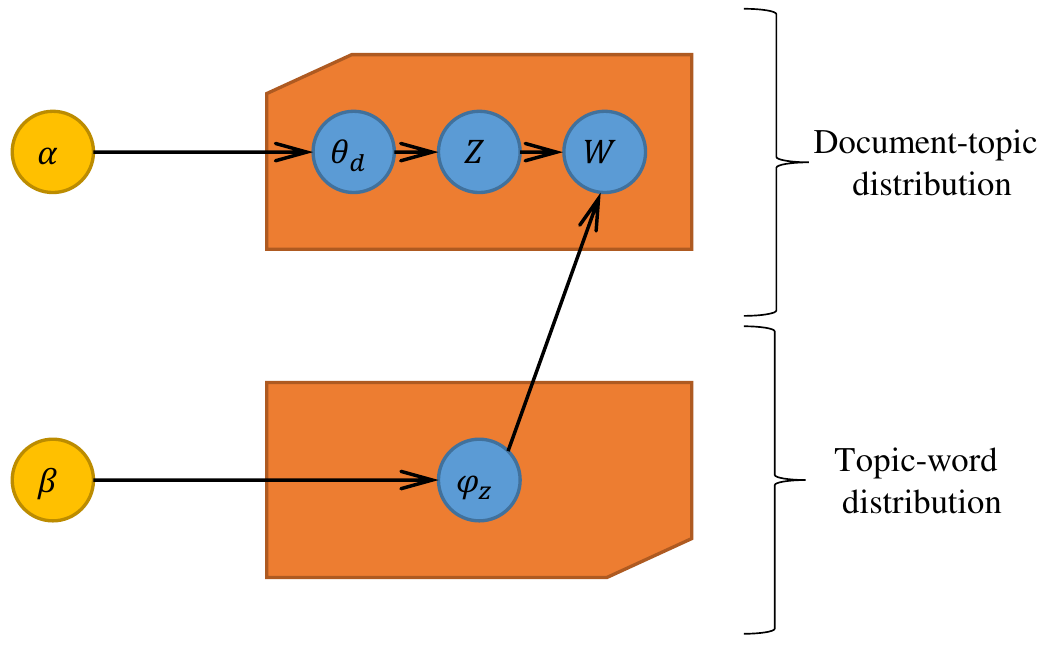}
\caption{Parameter relationship of LDA model.}
\label{fig_sim}
\end{figure}
 As shown in the Fig. 5, the Bayesian probability model of LDA can be divided into two parts. In the first part, the topic distribution ${\theta_d}$ of document $d$ is generated by sampling from the Dirichlet distribution $\alpha$, which can be expressed as ${\theta_d}\sim Dir(\alpha)$.  The topic $Z$ of each word in document $d$ is generated from the polynomial distribution of the topic.
In the second part, the LDA topic model samples from the Dirichlet distribution $\beta$ to generate the word polynomial distribution ${\varphi_Z}$ of topic $Z$ which expressed as ${\theta_Z}\sim Dir(\beta)$ . The final word $W$ is generated from the word polynomial distribution ${\varphi_Z}$.
$\alpha$ represents the prior parameters of Dirichlet distribution of document topic and $\beta$ is the prior parameters of Dirichlet distribution of topic words. ${\theta_d}$ represents the topic distribution in document $d$ and $Z$ represents the corresponding topic set. ${\varphi_Z}$ represents the word components contained in topic $Z$.
According to the principle of LDA topic, a document can be obtained from the probability distribution of document topic.
That is to say, for document ${D_ (i,j)}$ (i.e. $j$-$th$ topic in topic set $i$) can be obtained by polynomial distribution ${D_ (i,j)}\sim Mult({\theta_ d})$. Topic ${Z_(i,j)}$ corresponding to the $j-th$ word in $i-th$ , can be obtained by polynomial distribution ${Z_(d,i)}\sim Mult({\varphi_Z}) $.
In the parameter setting of LDA topic model, the prior parameters $\alpha$ and $\beta$  of Dirichlet distribution are set by experience. The posteriori parameters of the polynomial distribution, ${\theta_ d}$  and  ${\varphi_Z}$ need to be estimated by calculating the corresponding posteriori probability distribution from the data in the known corpus. Therefore, this paper employ Gibbs sampling to calculate the posteriori parameters ${\theta_ d}$  and ${\varphi_Z}$.
Considering the real data is usually difficult to find out the corresponding accurate probability distribution. Therefore, approximate inference method is often used to randomly fit the real probability distribution by sampling and Gibbs sampling is based on this idea. Gibbs sampling samplings $m$  $n$-$dimensional$  data $[{X_i]}_{i=1}^{n}$ from a Joint probability distribution [22]. The vectors $X_i$ obtained by sampling are initialize randomly. Each sample $x_ i$ can be derived from the conditional probability distribution $P (x_ i^j |X_ i^j,...,X_ i^j,X_{i-1}^{j+1},...,X_ {i-1}^{n})$ and $x_i^j$ represents the value of the $j$-$th$ dimension of sample $x_i$. The sampling formula of Gibbs sampling is shown in Equation 23.
\begin{equation}
\label{eqn_example}
P({Z_i} = K|{Z_{i - 1}},W) \propto \frac{{\left( {n_{k - i}^{\left( i \right)} + {\beta _i}} \right)\left( {n_{d - i}^{\left( k \right)} + {\alpha _k}} \right)}}{{\left( {\mathop \sum \nolimits_{i - 1}^V n_{k - i}^{\left( i \right)} + {\beta _i}} \right)}}
\end{equation}	
after substituting the parameters $\alpha$ and $\beta$  of LDA topic model, the posterior probability distribution of LDA topic and vocabulary can be obtained as shown in Equation (24).
\begin{equation}
\label{eqn_example}
P\left( {z,w|\alpha ,\beta } \right) = \mathop \prod \limits_{z = 1}^T \frac{{\Delta \left( {{n_z} + \beta } \right)}}{{\Delta \beta }}{\rm{*}}\mathop \prod \limits_{d = 1}^D \frac{{\Delta \left( {{n_d} + \alpha } \right)}}{{\Delta \alpha }}
\end{equation}	
when Gibbs sampling algorithm converges, the document topic probability distribution ${\theta_ d}$  and word topic probability ${\varphi_Z}$ can be obtained as shown in Equations (25) and (26).
\begin{equation}
\label{eqn_example}
{\theta _{d,z}} = \frac{{n_d^Z + \alpha }}{{\mathop \sum \nolimits_i^V n_z^i + \beta }}
\end{equation}	
\begin{equation}
\label{eqn_example}
{\varphi _{z,i}} = \frac{{n_z^i + \beta }}{{\mathop \sum \nolimits_{i = 1}^V n_z^i + \beta }}
\end{equation}	

Finally,  the topic probability distribution of a document and the vocabulary probability distribution of each topic can be obtained, so as to realize the topic mining of the document.
\subsection{Topic-word model}
Word representation is the minimum granularity representation in text representation and word vector representation model word2vec is applied in various fields of NLP. However, the word vector generated by word2vec cannot solve the problem of polysemy in natural language. In word2vec, a word has only one word vector, but in fact, some words in natural language contain many different meanings and even some words in different contexts represent far different semantic information. For example, The word "apple" means a fruit in the context of food and an IT company in the context of mobile phones. Obviously, the word vector representation method of word2vec has some irrationality. On the basis of skip-gram model, Liu et al. proposed a model based on topical word embeddings (TWE) [23]. This model introduces topic vector while training word vector, which aims to achieve different word vector representation under different topics by topic vector. Each topic in the model is trained as a word and the model learns the topic embedding of topic $z_i$  and topic embedding of word $w_i$ separately. Then topic word embedding $<w_i,z_i>$ is trained according to topic embedding and topic embedding.
\begin{figure}[!t]
\centering
\includegraphics[width=3in]{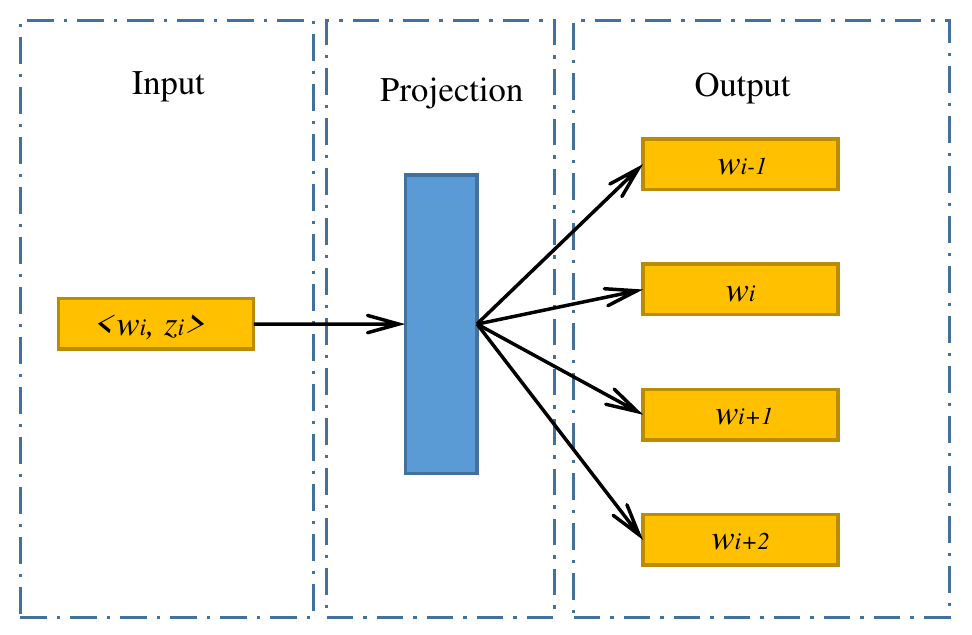}
\caption{The structure of TWE model.}
\label{fig_sim}
\end{figure}
\begin{figure*}[!t]
\centering
\includegraphics[width=6in]{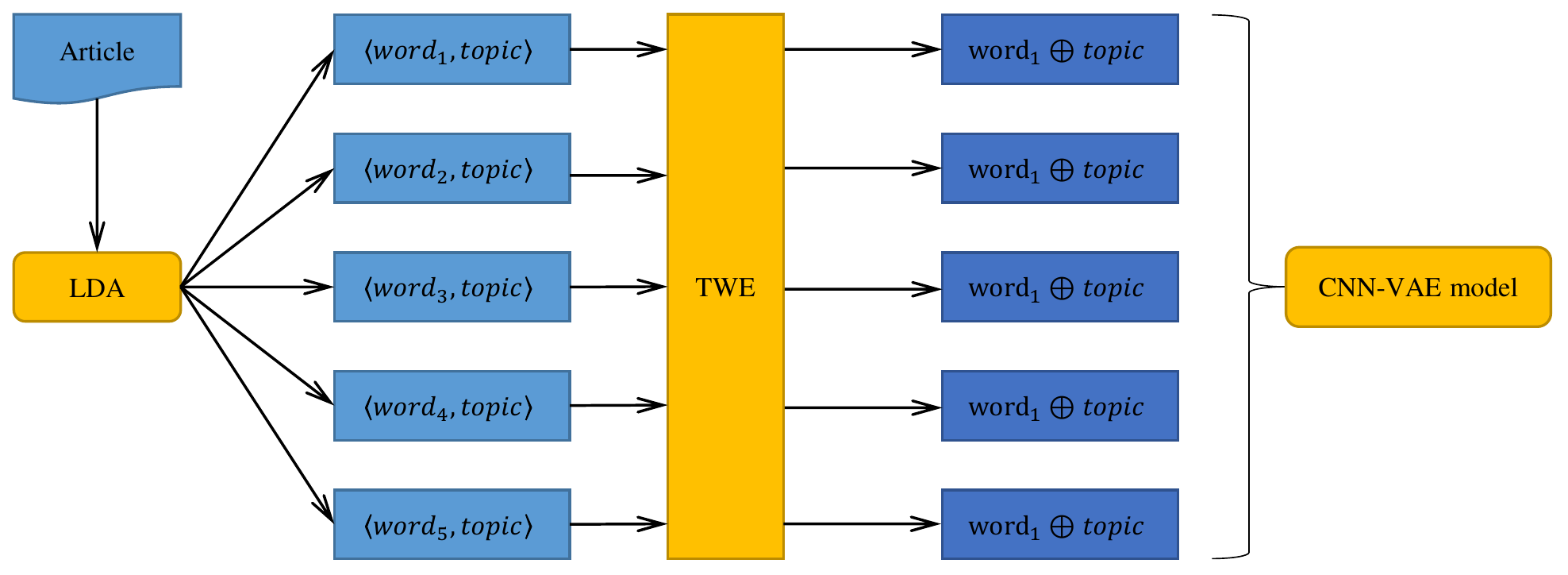}
\caption{Text representation model of CNN-AVE based on text semantics.}
\label{fig_sim}
\end{figure*}
TWE aims to learn the vector representation of words and topics at the same time and its structure is shown in the Fig. 6.
Compared with skip-gram using the central word $w_i$ in the sliding window to predict context, this model employs central word $w_i$ adding semantic information topic $z_i$ to predict context. This model aims to solve the problem of polysemy by training the word which have its corresponding vector under each topic. Topic word embedding of word $w$ in topic $z$ can be obtained by connecting word embedding with topic embedding as shown in Equation (27).
\begin{equation}
\label{eqn_example}
{w^z} = w \oplus z
\end{equation}	
where $\oplus$ is a cascading operation and the vector dimension of $w^z$ is twice that of $w$ or $z$.

The input of CNN-VAE model are the word vectors pre-trained by word2vec, which cannot solve polysemy. Therefore, this paper employs topic words method to get the word vectors meeting the requirements as shown in Fig. 7. Firstly LDA model is utilized to train every word in the text to get its corresponding topic number, that is, to convert the word in the text into $<word: topic$ $  number>$. Next, the topic vector $z$ and the word vector $w$ are trained based on the topic word vector model. Finally, the input vector $w^z$ of CNN-VAE is generated according to $<word: topic$ $  number>$.
\begin{figure}[!t]
\centering
\includegraphics[width=3.4in]{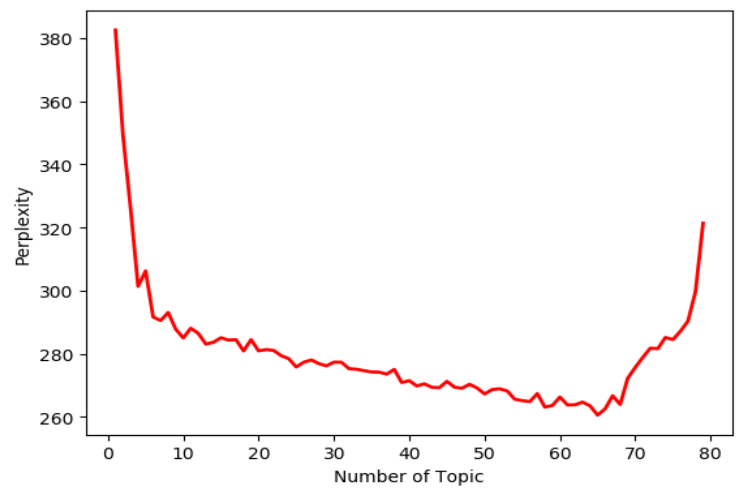}
\caption{ The relationship between the degree of perplexity and the number of topics.}
\label{fig_sim}
\end{figure}
\begin{figure}[!t]
\centering
\includegraphics[width=3.4in]{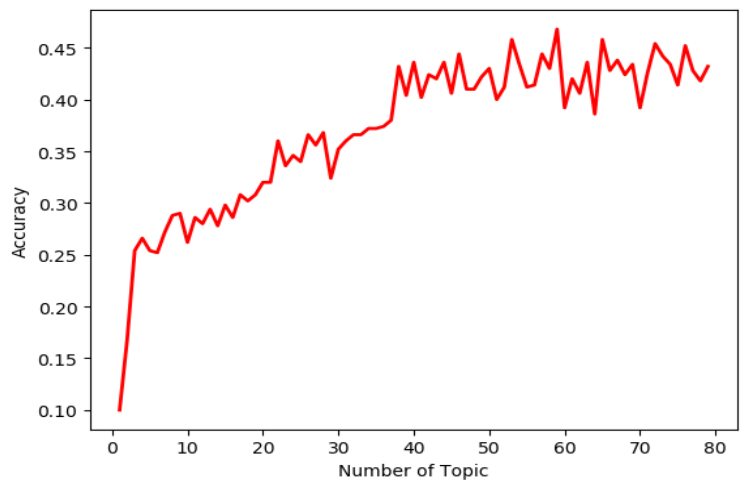}
\caption{ The relationship between the accuracy and the number of topics.}
\label{fig_sim}
\end{figure}

The number of topics needs to be determined before LDA topic model is used to generate topics. In this paper, two methods are used to determine the number of topics. (1) Calculate the perplexity of different  number of  topics.
(2) The probability of each document under all topics is calculated and transformed into its corresponding document vector, which is used to input into SVM classifier to compare the classification accuracy in different topics. The final number of topics is determine by comparing the above two methods.
The degree of perplexity is inversely proportional to the fitting ability of the model and the smaller the degree of perplexity of the model which has a better fitting effect on the text. In the LDA topic model, its perplexity is calculated as shown in Equation (28).
\begin{equation}
\label{eqn_example}
preplexity\left( D \right) = exp\left\{ { - \frac{{\mathop \sum \nolimits_{d = 1}^M \log p\left( {{w_d}} \right)}}{{\mathop \sum \nolimits_{d = 1}^M {N_d}}}} \right\}
\end{equation}	
where $D$ is the test set in the corpus, $N_d$ is the number of words in each document, $w_ d$ is the word in document $d$, $P (w_ d)$  is the probability of generating word $w_d$ in the document.

In this paper, the LDA model is trained by the training set and 1\% of them are randomly selected as the test set to calculate the degree of perplexity. The number of topics is set from 1 to 80. The relationship between the degree of confusion and the number of topics is shown in Fig. 8. and the relationship between the number of topics and the classification accuracy is shown in Fig. 9.
In Fig. 8, the degree of perplexity reaches the minimum value when the number of topics is 65, which indicates that when the number of topics is set to 65, the model is the best for data fitting and prediction.
In Fig. 9, when the number of topics is less than 45, the number of topics is proportional to the accuracy. When the number of topics is higher than 45, the accuracy will fluctuate in a certain range. Therefore, it can be concluded that after the number of settings exceeds 45, the model is stable for data fitting and prediction .
In conclusion,  the number of topics is set to 65 and all words in the corpus are transformed into the form of $<word ID: topic ID>$ after LDA training.
\begin{table}
\makeatletter\def\@captype{table}\makeatother
\caption{Cnews dataset.}
\label{tab:1}       
\centering
\setlength{\tabcolsep}{3mm}
\begin{tabular}{c c c c}
\hline\noalign{\smallskip}
Category	&Training set & Testing set & Validation set\\
\noalign{\smallskip}\hline\noalign{\smallskip}
Sports	& 5000 &1000 &500\\
Entertainment	& 5000 &1000 &500\\
Home furnishing	& 5000 &1000 &500\\
House property	& 5000 &1000 &500\\
Education	& 5000 &1000 &500\\
Fashion	& 5000 &1000 &500\\
Current affairs	& 5000 &1000 &500\\
Games	& 5000 &1000 &500\\
Science and technology	& 5000 &1000 &500\\
Finance and economics	& 5000 &1000 &500\\
Total &50000&10000&5000\\
\noalign{\smallskip}\hline
\end{tabular}
\end{table}
\begin{figure}[!t]
\centering
\includegraphics[width=3.4in]{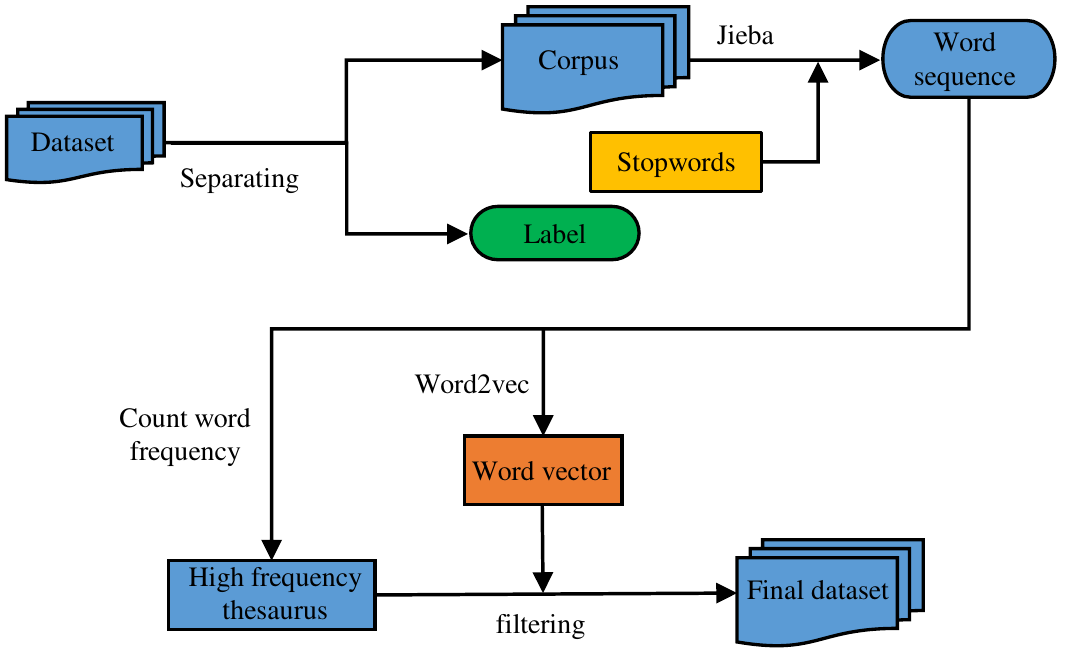}
\caption{Data preprocessing.}
\label{fig_sim}
\end{figure}
\section{Experiments}
\subsection{Dataset and Preprocessing}
This paper uses the open dataset Cnews (http://rss.sina.com.cn/news/) to verify the performance of the model, which is generated by filtering the historical data from RSS subscription channel of Sina News from 2005 to 2011. The dataset contains 10 categories of news, namely sports, entertainment, home furnishing, real estate, education, fashion, current affairs, games, technology and finance. There are 65000 text data in the dataset, which are divided into 50000 training data, 10000 testing data and 5000 validation data as shown in Table II.
Each row in the Cnews dataset represents an article and the beginning of each row corresponds to the news category of the article. Therefore, this paper extracts the tag information (i.e. news category) in the dataset and the dataset is divided into tag and corpus. Moreover, The Jieba segmentation is applied to process the article and the stoppage word is employed to filter out some words to improve the performance and computational efficiency of the subsequent experiments. Next, this paper makes word frequency statistics for the words appearing in the corpus and rank them according to the word frequency from large to small. Finally, the final data set can be obtained by eliminating the words that appear less frequently than the first 10000. The data preprocessing is shown as Fig. 10.

\subsection{Evaluating Metrics}
This paper employs four evaluation metrics: accuracy, recall, precision and F-score in order to verify the performance of the proposed model.
The formulas of four indictors are shown in Equations 29, 30, 31, and 32 respectively.
\begin{equation}
\label{eqn_example}
Accuracy = \frac{{TP + TN}}{{TP + FP + FN + TN}}
\end{equation}	
\begin{equation}
\label{eqn_example}
Recall = \frac{{TP}}{{TP + FN}}
\end{equation}	
\begin{equation}
\label{eqn_example}
Precision = \frac{{TP}}{{TP + FP}}
\end{equation}	
\begin{equation}
\label{eqn_example}
F1 - score = 2 \times \frac{{precision \times recall}}{{precison + recall}}
\end{equation}	
where the true positive (TP) indicates that the prediction category is positive and the actual category is  positive. The false positive (FP) indicates that the prediction category is positive and the actual category is negative. The false negative (FN) indicates that the prediction category is negative and the real category is positive. The true negative (TN) indicates that the prediction category is negative and the actual category is negative.
\subsection{Experimental Results and Analysis}
This paper compares the proposed model with two unsupervised text representation methods that convert word vectors into document vectors. The control models are shown as follows.
(1) w2v-avg: This model averages all word vectors of each document to get the final representation. This experiment sets the word vector dimension of training to 128 and get the text representation vector dimension to 128.
(2) CNN-AE: An AE model based on CNN, which can get the text representation via using word vectors obtained by CNN as the input of encoder.
The experimental results are shown in tables  II, III and IV.
\begin{table}
\makeatletter\def\@captype{table}\makeatother
\caption{Performance comparison among w2v-avg,  CNN-AE and CNN-VAE under KNN classification algorithm}
\label{tab:1}       
\setlength{\tabcolsep}{1.5mm}
\centering
\begin{tabular}{ccccc}
\hline\noalign{\smallskip}
        & Accuracy      & Recall        &Precision      & F1-Score\\
\noalign{\smallskip}\hline\noalign{\smallskip}
w2v-avg	&	87.32$\pm$7.40	&77.02$\pm$7.31	    &80.82$\pm$7.61	    &77.28$\pm$7.38\\
CNN-AE	&	92.94$\pm$5.06	&86.44$\pm$10.69	&89.56$\pm$11.12	&86.66$\pm$10.70\\
CNN-VAE	&	\textbf{94.87$\pm$4.61}	&\textbf{91.81$\pm$7.46}	    &\textbf{91.87$\pm$9.12}	    &\textbf{90.87$\pm$8.08}\\
\noalign{\smallskip}\hline
\end{tabular}
\end{table}
\begin{table}
\makeatletter\def\@captype{table}\makeatother
\caption{Performance comparison among w2v-avg,  CNN-AE and CNN-VAE under RF classification algorithm}
\label{tab:1}       
\centering
\setlength{\tabcolsep}{1.5mm}
\begin{tabular}{ccccc}
\hline\noalign{\smallskip}
        & Accuracy      & Recall        &Precision      & F1-Score\\
\noalign{\smallskip}\hline\noalign{\smallskip}
w2v-avg		&85.48$\pm$0.93    &66.23$\pm$2.27	&74.84$\pm$4.25     &68.27$\pm$2.73\\
CNN-AE		&92.43$\pm$0.66    &80.26$\pm$1.42	&92.64$\pm$3.18     &83.97$\pm$1.85\\
CNN-VAE		&\textbf{94.98$\pm$0.65}    &\textbf{92.54$\pm$1.67}	&\textbf{95.29$\pm$1.00}     &\textbf{93.37$\pm$1.22}\\
\noalign{\smallskip}\hline
\end{tabular}
\end{table}
\begin{table}
\makeatletter\def\@captype{table}\makeatother
\caption{Performance comparison among w2v-avg,  CNN-AE and CNN-VAE under SVM classification algorithm}
\label{tab:1}       
\centering
\setlength{\tabcolsep}{1.5mm}
\begin{tabular}{ccccc}
\hline\noalign{\smallskip}
        & Accuracy      & Recall        &Precision      & F1-Score\\
\noalign{\smallskip}\hline\noalign{\smallskip}
w2v-avg		&86.51$\pm$3.18	&87.61$\pm$3.08	&88.68$\pm$2.83	&87.58$\pm$3.27\\
CNN-AE		&90.00$\pm$1.64	&90.94$\pm$1.48	&90.10$\pm$2.98	&90.64$\pm$1.55\\
CNN-VAE		&\textbf{93.30$\pm$2.63}	&\textbf{93.90$\pm$2.08}	&\textbf{94.29$\pm$1.90}	&\textbf{93.83$\pm$1.99}\\
\noalign{\smallskip}\hline
\end{tabular}
\end{table}
In Table III, IV and V, the performance of CNN-AE is better than that of w2v-avg. Therefore, CNN is conducive to extracting the local semantic features of the components among words. In addition, when the text representation dimension is 128, the proposed model is superior to CNN-AE and w2v-avg in KNN, RF and SVM classifiers. That is to say, if AE is improved to VAE, the text feature space can better fit the Gaussian distribution, which makes the semantic information conform to the real distribution. In conclusion, the VAE text feature representation model based on CNN is effective and feasible.

\begin{table}
\makeatletter\def\@captype{table}\makeatother
\caption{Performance comparison between word2vec+CNN-VAE and TWE+CNN-VAE under KNN classification algorithm}
\label{tab:1}       
\setlength{\tabcolsep}{0.8mm}
\centering
\begin{tabular}{ccccc}
\hline\noalign{\smallskip}
        & Accuracy      & Recall        &Precision      & F1-Score\\
\noalign{\smallskip}\hline\noalign{\smallskip}
word2vec+CNN-VAE	&	\textbf{94.87$\pm$4.61}	&91.81$\pm$7.46	    &91.87$\pm$9.12    &90.87$\pm$8.08\\
TWE+CNN-VAE	&	94.42$\pm$5.27	&\textbf{91.83$\pm$7.45}	&\textbf{93.34$\pm$7.75}	&\textbf{91.47$\pm$7.24}\\
\noalign{\smallskip}\hline
\end{tabular}
\end{table}
\begin{table}
\makeatletter\def\@captype{table}\makeatother
\caption{Performance comparison between word2vec+CNN-VAE and TWE+CNN-VAE under RF classification algorithm}
\label{tab:1}       
\centering
\setlength{\tabcolsep}{0.8mm}
\begin{tabular}{ccccc}
\hline\noalign{\smallskip}
        & Accuracy      & Recall        &Precision      & F1-Score\\
\noalign{\smallskip}\hline\noalign{\smallskip}
word2vec+CNN-VAE		&94.98$\pm$0.65    &92.54$\pm$1.67	&95.29$\pm$1.00     &93.37$\pm$1.22\\
TWE+CNN-VAE		&\textbf{96.85$\pm$0.51}    &\textbf{92.76$\pm$1.09}	&\textbf{96.65$\pm$0.46}     &\textbf{94.92$\pm$0.79}\\
\noalign{\smallskip}\hline
\end{tabular}
\end{table}
\begin{table}
\makeatletter\def\@captype{table}\makeatother
\caption{Performance comparison between word2vec+CNN-VAE and TWE+CNN-VAE under SVM classification algorithm}
\label{tab:1}       
\centering
\setlength{\tabcolsep}{0.8mm}
\begin{tabular}{ccccc}
\hline\noalign{\smallskip}
        & Accuracy      & Recall        &Precision      & F1-Score\\
\noalign{\smallskip}\hline\noalign{\smallskip}
word2vec+CNN-VAE		&93.30$\pm$2.63	&\textbf{93.90$\pm$2.08}	&\textbf{94.29$\pm$1.90} &93.83$\pm$1.99\\
TWE+CNN-VAE		&\textbf{93.49$\pm$3.53}	&93.87$\pm$3.21	&94.23$\pm$2.77	&\textbf{94.05$\pm$3.20}\\

\noalign{\smallskip}\hline
\end{tabular}
\end{table}

In order to verify the performance of CNN-VAE model based on the two, the output of word2vec and the output of TWE are compared as the input of CNN-VAE. The experimental results are shown in Table VI, VII and VIII. Table VI shows the classification results of the two models as pre- training under KNN classification algorithm. TWE+CNN-VAE is higher than word2vec+CNN-VAE in Recall, Precision and F-score, only 0.45\% lower in accuracy. Table VII show that the results of the four evaluation metrics are higher than those of word2vec+CNN-VAE in the RF classifiers. Table VIII show that the results of accuracy and F1-Score are higher than those of word2vec+CNN-VAE in the SVM classifiers.  The experimental results show that the combination of topic information can solve the problem of polysemy to a certain extent.

\section{Conclusion}
Text feature representation plays a significant role in the field of NLP as the first step of machine recognition of natural language, which extracts the features of the text to express semantic information contained in the text. With the improvement of hardware performance, deep learning and neural network, text representation has leap forward progress. At the same time, the increasing number of Internet users produces a large number of unstructured text data. How to employ NLP technology to analyze and process huge text data has become the current research hotspots. Therefore, this paper proposes a text feature representation model based on CNN-VAE, which is feature representation method from word vector to text vector. On the one hand, the model applies CNN to extract the features of the text vector as the text representation, so as to better extract the semantics among words. On the other hand, the model uses VAE instead of AE to make the text feature space better fit the Gaussian distribution, so that the semantic information is more consistent with the real distribution. In addition, the output of the improved word2vec model is employed as the input of the proposed model to distinguish different meanings of the same word in different contexts.
 In order to verify the performance of the model, the proposed model is compared with w2v-avg and CNN-AE. Experimental results show that the proposed model has better performance in KNN, RF and SVM classification.
There are four levels of language structure including document, paragraph, sentence and word in NLP. This paper only involves two different levels of text vector representation and attention should be paid to different levels of language structure in the future.

%

\begin{IEEEbiography}[{\includegraphics[width=1in,height=1.25in,clip,keepaspectratio]{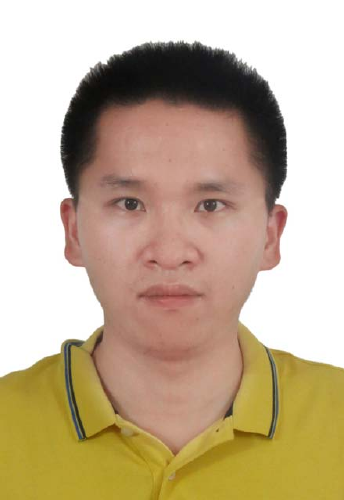}}]{Genggeng Liu} received the B.S. degree in Computer Science from Fuzhou University, Fuzhou, China, in 2009, and the Ph.D. degree in Applied Mathematics from Fuzhou University in 2015. He is currently an associate professor with the College of Mathematics and Computer Science at Fuzhou University. His research interests include computational intelligence and its application.
\end{IEEEbiography}
\begin{IEEEbiography}[{\includegraphics[width=1in,height=1.25in,clip,keepaspectratio]{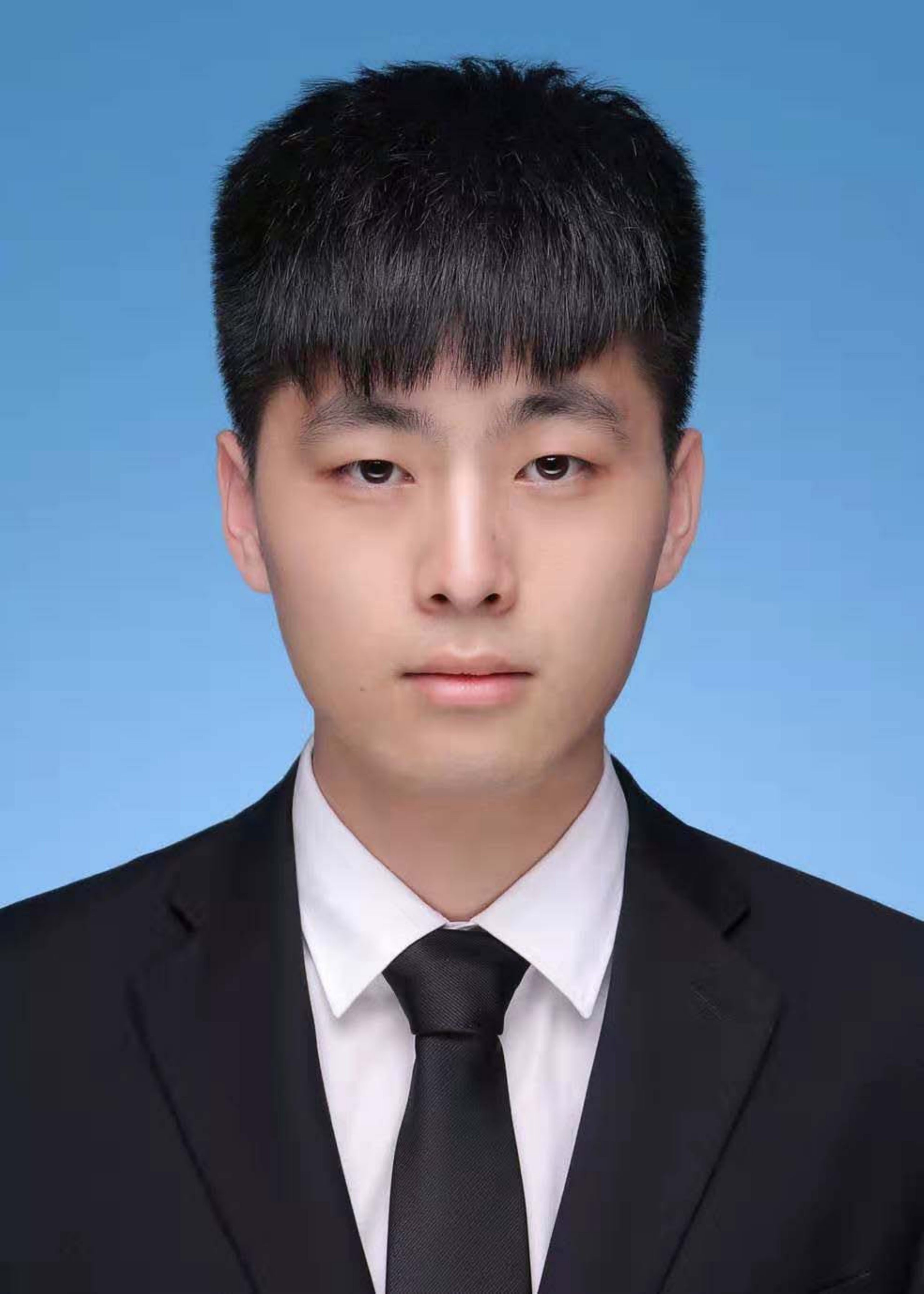}}]{Canyang Guo}  is current a master student at College of Mathematics and Computer Sciences, Fuzhou University, Fuzhou, China. His research interests include machine learning and big data.
\end{IEEEbiography}
\begin{IEEEbiography}[{\includegraphics[width=1in,height=1.25in,clip,keepaspectratio]{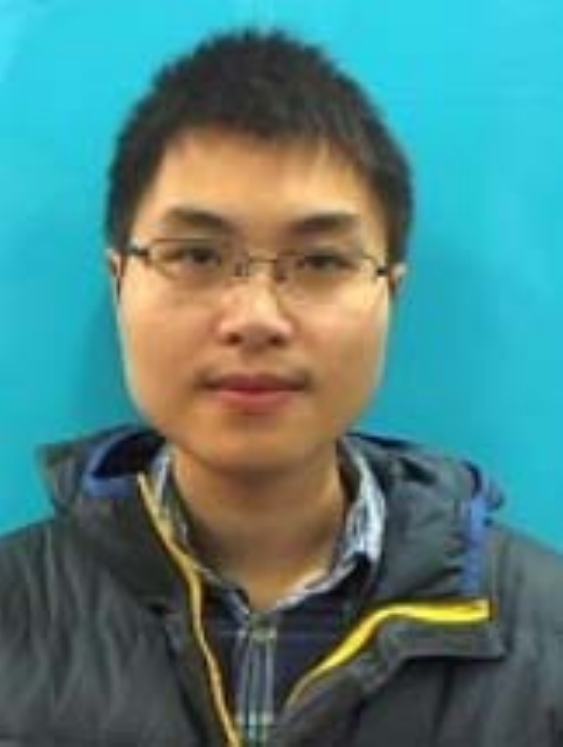}}]{Lin Xie} is current a master student at College of Mathematics and Computer Sciences, Fuzhou University, Fuzhou, China. His research interests include machine learning and text representation.
\end{IEEEbiography}
\begin{IEEEbiography}[{\includegraphics[width=1in,height=1.25in,clip,keepaspectratio]{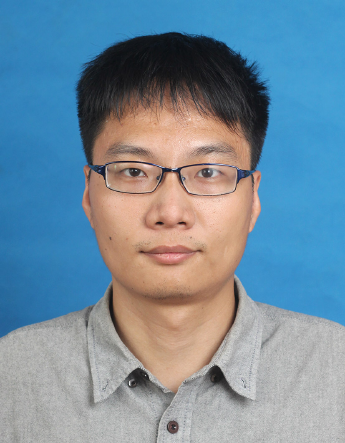}}]{Wenxi Liu} is an Associate Professor in the College of Mathematics and Computer Science, Fuzhou University. He obtained his Ph.D degree from City University of Hong Kong. His research interests include computer vision, robot vision, and image processing.
\end{IEEEbiography}

\begin{IEEEbiography}[{\includegraphics[width=1.2in,
height=1.0in,clip,keepaspectratio]{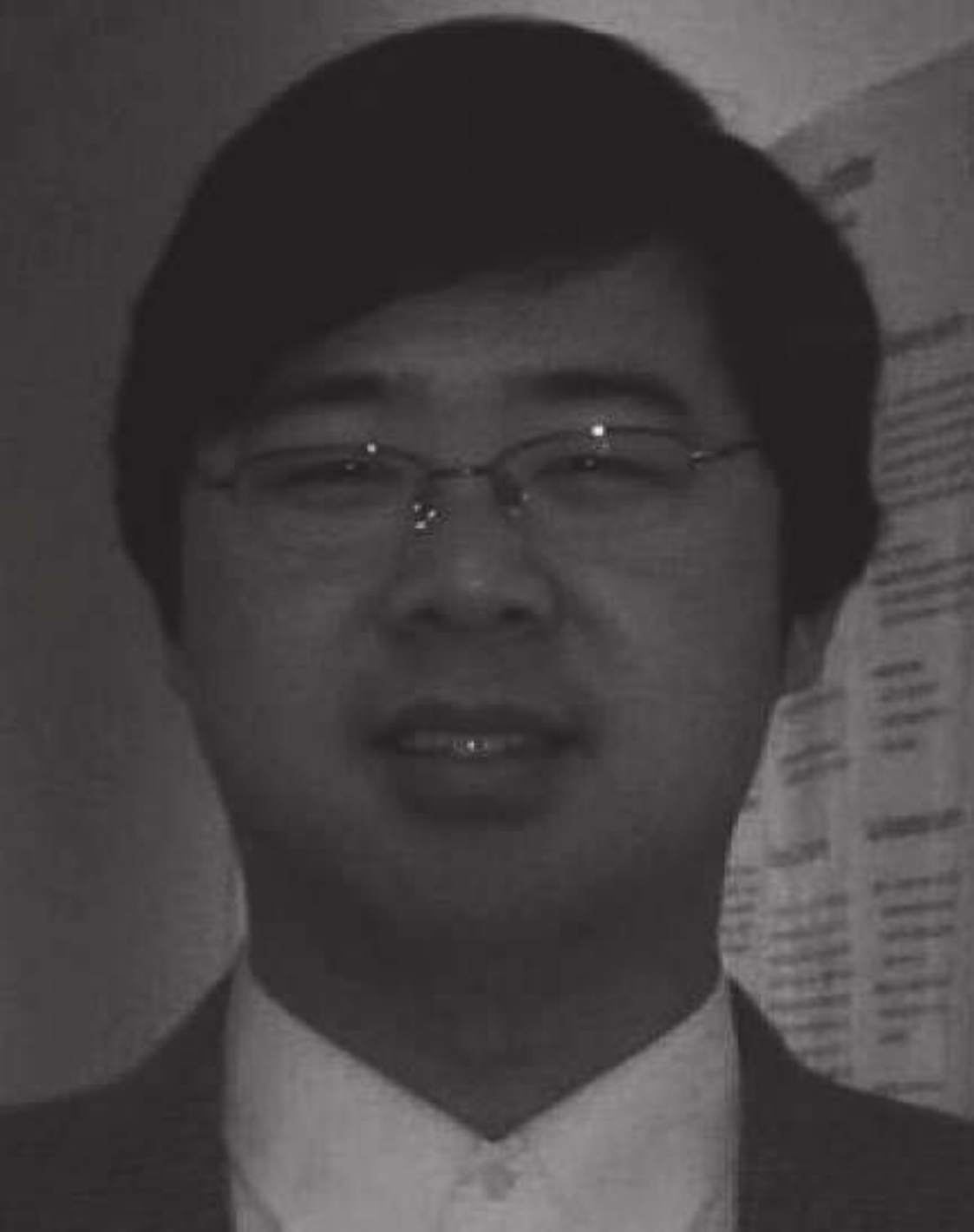}}]{Naixue Xiong}
received the B.E.
degree in computer science from the Hubei University
of Technology, Wuhan, China, in 2001,
the M.E. degree in computer science from Central
China Normal University, Wuhan, China, in
2004, and Ph.D. degrees in software engineering
from Wuhan University, Wuhan, China, in
2007, and in dependable networks from the
Japan Advanced Institute of Science and Technology,
Nomi, Japan, in 2008.
He is current a Full Professor at the Department
of Business and Computer Science, Southwestern Oklahoma State
University, Weatherford, OK, USA. His research interests include cloud
computing, security and dependability, parallel and distributed computing,
networks, and optimization theory.
Dr. Xiong serves as Editor-in-Chief, Associate Editor or Editor Member,
and Guest Editor for more than ten international journals including
as an Associate Editor of the IEEE TRANSACTIONS ON SYSTEMS, MAN
\& CYBERNETICS: SYSTEMS, and Editor-in-Chief of the Journal of Parallel
and Cloud Computing, the Sensor Journal, WINET, and MONET.
\end{IEEEbiography}

\begin{IEEEbiography}[{\includegraphics[width=1.2in,
height=1.1in,clip,keepaspectratio]{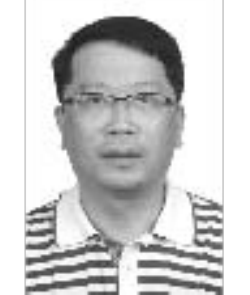}}]{Guolong Chen}
received the B.S. and M.S degrees in Computational Mathematics from
Fuzhou University, Fuzhou, China, in 1987 and 1992, respectively,
and the Ph.D degree in Computer Science from Xi'an Jiaotong
University, Xi'an, China, in 2002. He is a professor with the
College of Mathematics and Computer Science at Fuzhou University.
His research interests include computation intelligence, computer
networks, information security, etc.
\end{IEEEbiography}






\end{document}